\pdfoutput=1

\documentclass[11pt]{article}

\usepackage[preprint]{acl}

\usepackage{times}
\usepackage{latexsym}

\usepackage[T1]{fontenc}

\usepackage[utf8]{inputenc}

\usepackage{microtype}

\usepackage{inconsolata}

\usepackage{graphicx}

\usepackage{tabularx}
\usepackage{amsmath}
\usepackage{graphicx} 
\usepackage{enumitem}
\usepackage{booktabs}
\usepackage{diagbox}
\usepackage{float}
\usepackage{multirow}
\usepackage{booktabs}
\usepackage{colortbl} 
\definecolor{deepred}{rgb}{0.7, 0.0, 0.0}
\definecolor{deepblue}{rgb}{0.36, 0.44, 0.58}
\definecolor{deepgreen}{rgb}{0.38, 0.49, 0.38}
\title{CEM: A Data-Efficient Method for Large Language Models to Continue Evolving From Mistakes}

  \author {
    Haokun Zhao\textsuperscript{\rm 1},
    Jinyi Han\textsuperscript{\rm 2},
    Jie Shi\textsuperscript{\rm 1},
    Chengyu Du \textsuperscript{\rm 1},
    Jiaqing Liang\textsuperscript{\rm 3 \thanks{Corresponding Author}},
    Yanghua Xiao \textsuperscript{\rm 1,2 }
    \\
    Shanghai Key Laboratory of Data Science, School of Computer Science, Fudan University\textsuperscript{\rm 1}\\
    Shanghai Institute of Artificial Intelligence for Education, East China Normal University\textsuperscript{\rm 2}\\
    Shanghai Key Laboratory of Data Science, School of Data Science, Fudan University\textsuperscript{\rm 3}\\
   \texttt{\{hkzhao23,jshi22\}@m.fudan.edu.cn}, 
    \texttt{\{haixiahan03,cydu2024\}@gmail.com}, \\
    \texttt{\{liangjiaqing,shawyh\}@fudan.edu.cn}
}

\begin{document}
\maketitle
\begin{abstract}


As world knowledge advances and new task schemas emerge, Continual Learning (CL) becomes essential for keeping Large Language Models (LLMs) current and addressing their shortcomings. This process typically involves continual instruction tuning (CIT) and continual pre-training (CPT) to enable these models to adapt to novel tasks and acquire supplementary knowledge. However, collecting sufficient CPT data and efficiently bridging knowledge gaps remain significant challenges. Inspired by the `summarizing mistakes' strategy, we propose the \textbf{\underline{C}ontinue \underline{E}volving from \underline{M}istakes (CEM)} method, a data-efficient approach aiming to collect CPT data and continually improve LLMs' performance through iterative evaluation and supplementation with mistake-relevant knowledge. To further optimize data usage and mitigate forgetting, we introduce a novel training paradigm that combines CIT and CPT. Experiments show that CEM substantially enhances multiple models’ performance on both in-domain and out-of-domain QA tasks, achieving gains of up to 29.63\%. Code and datasets are available on GitHub \footnote{https://anonymous.4open.science/r/cem-BB25}.

\end{abstract}
\section{Introduction}

With the exponential growth of data and model sizes, Large Language Models (LLMs) have demonstrated superior performance across numerous downstream tasks \cite{chang2023survey,wei2022finetuned,peng2023instruction}. 
However, in real-world applications, the continual emergence of new knowledge and evolving task requirements necessitate ongoing updates and task-specific adaptations for LLMs. Without these, models risk hallucinating due to their limited knowledge boundaries and may cause task misalignment \cite{huang2023survey}.
Additionally, addressing and correcting the shortcomings and errors exposed during practical use is crucial. 
Therefore, Continual Instruction Tuning (CIT) and Continual Pre-training (CPT) are proposed as the primary methods of \textbf{Continual Learning (CL)} to align LLMs with evolving knowledge and tasks, and to improve their shortcomings \cite{wu2024continual}.

\begin{figure}[htb]
\small
  \centering
  \includegraphics[width=0.5\textwidth]{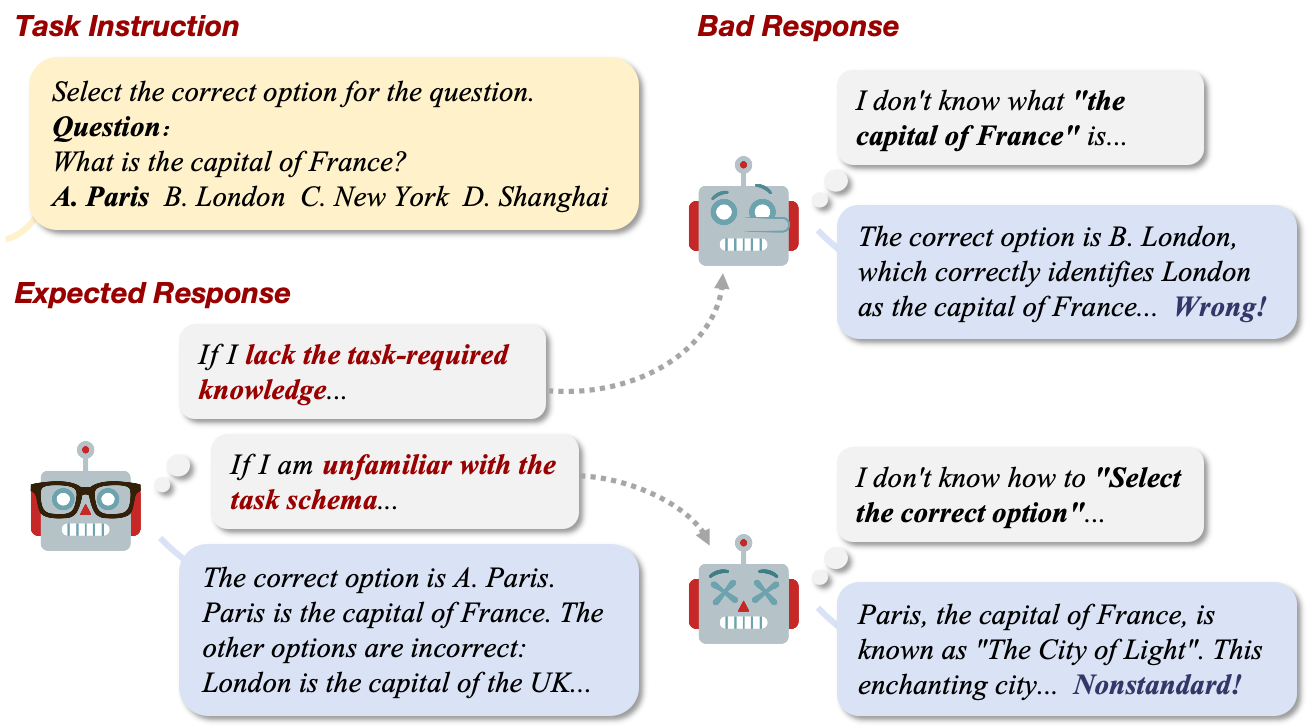}
  \caption{Two potential triggers for poor model performance: (1) \textbf{Task Schema Unfamiliarity}, and (2) \textbf{Lack of Task-relevant Knowledge}. Unfamiliarity with the task schema can cause deviations from expected interaction styles, while insufficient task knowledge may lead to hallucinations. Instruction tuning has been shown to be effective for addressing the former, but poor for the latter \cite{gekhman2024doesfinetuningllmsnew,zhou2023lima}.}
  \label{fig:introbig}
\end{figure}

Although CIT requires less data and poses a lower overfitting risk than CPT, it is less effective at injecting new knowledge. As \citet{zhou2023lima} highlight, instruction tuning is a superficial alignment focused on interaction styles (\textit{i.e.,} task schema) and is not well-suited for large-scale knowledge supplementation. This is because scaling up the necessary task-specific instruction data significantly increases complexity in terms of data selection, careful prompt design, and ensuring coverage across extensive knowledge domains. Consequently, CIT cannot address all issues shown in Figure \ref{fig:introbig}, where LLM underperformance may arise from a \textbf{lack of requisite knowledge} rather than \textbf{unfamiliarity with the task schema} \cite{ren2023investigating,zhang2023large}. In contrast, CPT better addresses such knowledge deficits.


However, using CPT to supplement the model with lacking knowledge still presents several challenges. First, \textbf{collecting CPT data} is challenging as identifying the lacking knowledge a priori while ensuring a sufficiently large data size is difficult. Blind data collection and insufficient data volume may introduce irrelevant or redundant information, weaken generalization, and increase overfitting risks \cite{zhang2024rtuning,gekhman2024does}.
Second, \textbf{efficiently utilizing CPT data and mitigating degradation} are difficult. As \citet{jiang2024instructiontuned} note, even at minimal perplexity, not all knowledge can be extracted—a `perplexity curse' that limits effective data utilization. Moreover, compared to supervised fine-tuning, the larger parameter shifts in CPT increase the risk of performance degradation.


To address these challenges, common data collection strategies rely on the latest Wikipedia snapshots or new domain academic papers \cite{cossu2022continual, jin2022lifelong}. However, these methods lack focus and efficiency and fail to dynamically update model deficiencies. Additionally, recent work has explored using highly relevant QA pairs for CIT before CPT \cite{jiang2024instructiontuned} or converting raw CPT corpora into CIT instructions (\textit{e.g.,} reading comprehension instructions) \cite{cheng2024adapting}, thereby reducing degradation and enhancing knowledge absorption. Nevertheless, their dependence on advanced LLMs and the complexity of mining patterns limit their practical applicability.



In this paper, we propose the \textbf{\underline{C}ontinue} \textbf{\underline{E}volving} \textbf{from} \textbf{\underline{M}istakes (CEM)} method, which: (1) \textbf{introduces a practical pipeline to collect CPT data} derived from model mistakes, and (2) \textbf{employs a novel training paradigm that integrates CIT and CPT} to efficiently utilize these data. Together, these innovations enable continual model improvement while mitigating performance degradation.

\textbf{To enhance the focus and efficiency of data collection}, we draw inspiration from the human learning skill of `summarizing mistakes'. We argue that a model's mistakes in `exams' reflect its inherent knowledge deficiencies. Accordingly, we collect background information from the internet related to these identified mistakes, directly addressing the specific knowledge deficiencies they reveal.
\textbf{To further optimize CPT data utilization and mitigate forgetting}, we construct training sets integrating CIT and CPT data in parallel. Coupled with methods to mitigate catastrophic forgetting (\textit{e.g.,} Random Replay) \cite{li2022large}, CEM supports iterative, continual model evolution. Across extensive experiments, CEM yields substantial gains for multiple models, increasing accuracy by up to 29.63\%.

\section{Related Work}

\textbf{Continual Pre-training (CPT)}

Continual pre-training involves regularly updating LLMs with the latest facts or adapting them to specialized domains \cite{wu2024continual}. To continually pre-train LLMs and update facts, researchers utilize dynamic datasets to assimilate real-time data from various sources such as news feeds \cite{sun2019ernie}, scholarly articles \cite{cossu2022continual}, and social media \cite{cossu2022continual}.

Continual pre-training enhances domain knowledge through two approaches: (1) domain-incremental pre-training, which accumulates knowledge across multiple domains, and (2) domain-specific continual learning, which refines a general model into a domain expert by training on domain-specific datasets and tasks. In domain-incremental pre-training, \citet{cossu2022continual} continually pre-trained LLMs on new multimodal data streams, preparing them for various asks. \citet{ke2023continual} introduced a soft-masking mechanism to update LLMs with domain corpora, aiming to boost performance while preserving general knowledge. Notably, there has been research on continual pre-training for specific domains such as finance \cite{xie2023efficient} and e-commerce \cite{ma2023ecomgptct}.

\textbf{Continual Instruction Tuning (CIT)}

Continual Instruction Tuning involves continually fine-tuning LLMs to learn how to follow instructions and transfer knowledge for future tasks \cite{zhang2023citb}. It can be broadly classified into two categories: (1) task-incremental CIT and (2) domain-incremental CIT \cite{wu2024continual}.

Task-incremental CIT aims to continually fine-tune LLMs on new task-specific instructions. While a straightforward approach involves generating new instruction-tuning data and fine-tuning LLMs directly \cite{wang2023trace}, this method can lead to catastrophic forgetting of previously learned knowledge and skills \cite{kotha2024understanding}.

Domain-incremental CIT aims to continually fine-tune LLMs on domain-specific instructions to acquire the knowledge required for solving tasks in new domains. TAPT \cite{gururangan-etal-2020-dont} adaptively tunes LLMs on domain-specific data from various domains, followed by evaluation of their text classification ability in each domain. AdaptLLM \cite{cheng2024adapting} enriches the raw training corpus with reading comprehension tasks, helping the model leverage domain-specific knowledge while enhancing prompting performance.
\section{Methods}
\setlength{\abovedisplayskip}{8pt} 
\setlength{\belowdisplayskip}{8pt}
\label{sec:methods}

\subsection{CEM Method}
We propose the CEM method, as illustrated in Figure \ref{fig:pipeline}. The main steps are as follows:

\begin{figure*}[htb]
  \centering
  \includegraphics[width=1.0\textwidth]{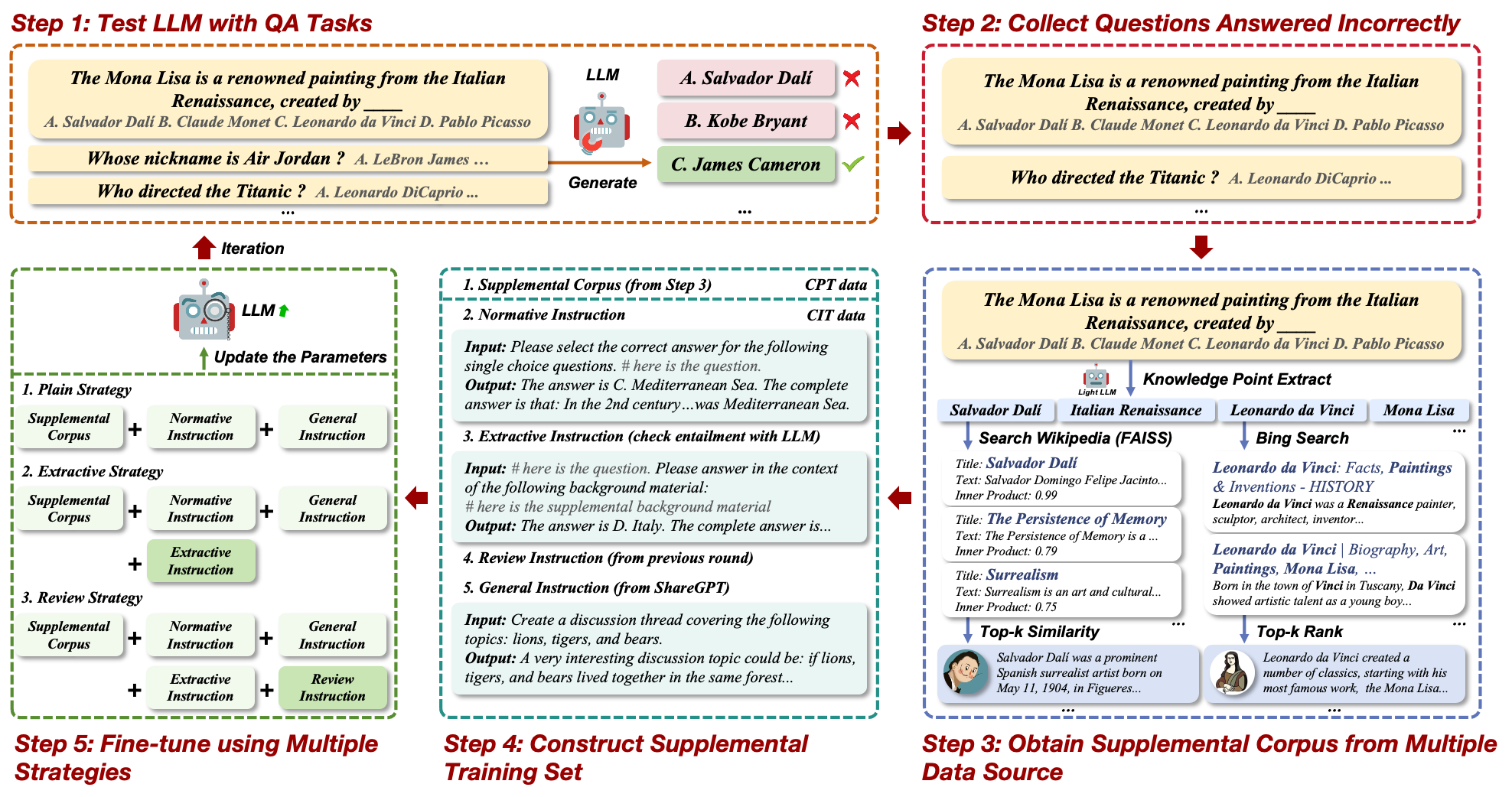}
  \caption{The pipeline of CEM method.}
  \label{fig:pipeline}
\end{figure*}

\textbf{Step 1\&2: Testing LLM and Collecting Incorrectly Answered Questions.} We use the LLM to answer questions from a QA dataset. By employing pattern matching to extract answers from the LLM's responses, we calculate the accuracy and collect the questions, along with their options, that the LLM answered incorrectly.

\textbf{Step 3: Obtaining Supplemental Corpus from Multiple Data Sources.} We first employ the lightweight LLM, \textbf{Qwen2-7B-instruct}\footnote{https://huggingface.co/Qwen/Qwen2-7B-Instruct} \cite{yang2024qwen2}, to identify knowledge points involved in questions erroneously answered by the tested LLM. For further details, see Appendix \ref{appendix:kp}. The extracted knowledge points are then utilized as keywords to aggregate supplemental corpora from diverse sources, including search engines and online encyclopedias.

(1) \emph{Search Engine:} Search engines are known for high relevance matching and diverse content. We search for extracted knowledge points using Bing and select the top-k results based on relevance, scraping the main textual content as a supplemental corpus. To prevent LLMs from simply memorizing answers, we employ a 13-gram matching to filter out any results that contain both the original question and its answer. Moreover, we use newspaper3k\footnote{https://github.com/codelucas/newspaper} and readability\footnote{https://github.com/buriy/python-readability} to exclude irrelevant and spam content.

(2) \emph{Online Encyclopedia:} Compared to search engines, online encyclopedias offer more credible and professional results, despite having lower hit rates. Since encyclopedias are typically organized by keyword entries, we employ the state-of-the-art massive text embedding model, \textbf{Conan-embedding-v1}\footnote{https://huggingface.co/TencentBAC/Conan-embedding-v1} \cite{li2024conanembeddinggeneraltextembedding}, to generate vector embeddings for Wikipedia entry titles. FAISS\footnote{https://github.com/facebookresearch/faiss} \cite{douze2024faiss} is then used to build an efficient vector index, enabling searches based on inner product similarity. By calculating the similarity between input keyword embeddings and the FAISS-indexed entry embeddings, we retrieve the top-k results that exceed a predefined similarity threshold, serving as the supplemental corpus. 

Appendix \ref{appendix:wikibing} presents the experimental parameters and processes.

\textbf{Step 4\&5: Constructing Supplemental Training Set and Fine-Tuning LLM.} The supplemental training sets are constructed separately based on the supplemental corpus collected from Wikipedia and Bing. They are then employed to fine-tune the LLM. The construction strategies of the training set will be explained in detail in Section \ref{sec:traningdataconstructionmethods}.

After completing one round of supplemental training, the model is retested on the QA dataset to evaluate its performance and gather a new set of incorrectly answered questions. This can serve as the first step for the next round of CEM. 

It should be noted that supplemental training should not be endless, as it may lead to catastrophic forgetting and diminishing returns in accuracy improvement (\textit{i.e.,} `perplexity curse'). Subsequent experiments explore the impact of the number of training rounds on model performance.

\subsection{Supplemental Training Data}
\label{sec:traningdataconstructionmethods}
\textbf{Definition of Data Type.} The construction of the supplemental training set involves the following five types of data.

(1) \emph{\underline{S}upplemental Corpus($C_{\underline{S}}$):} The supplemental corpus, obtained from multiple data sources, contains knowledge that the LLM lacks and is used as CPT data.  We employ \textbf{Qwen2-72B-Instruct}\footnote{https://huggingface.co/Qwen/Qwen2-72B-Instruct} \cite{yang2024qwen2} for entailment checking to verify whether the Supplemental Corpus contains information that helps the model answer correctly. Details are provided in the Appendix \ref{sec:relevant}.

Below are the four types of CIT data:

(2) \emph{\underline{N}ormative Instruction($I_{\underline{N}}$):}

To enhance the LLM's understanding of task schemas and formats, task-relevant instruction fine-tuning is commonly applied. For QA tasks, we sample questions from the training set and format them as shown in Table \ref{tab:normative} to build the Normative Instruction, enabling the LLM to answer multiple-choice questions in a standardized manner. Before supplemental training, we fine-tune the LLM using the Normative Instruction (\textit{i.e.,} pre-finetuning). A small portion of the Normative Instruction is incorporated during supplemental training to maintain the model's ability to provide normative answers.

(3) \emph{\underline{E}xtractive Instruction($I_{\underline{E}}$):} 
Inspired by the human learning process, where despite studying from the same textbooks, students who excel at extracting knowledge achieve superior grades, we believe that training models to effectively utilize materials is as crucial as acquiring them. 

To enhance the LLM's ability to capture and comprehend knowledge from the Supplemental Corpus, we restructured a subset of training set questions into an information-extraction question format, as illustrated in Table \ref{tab:extractive}. We employ Qwen2-72B-Instruct for entailment checking to filter the instructions. A minimal amount of Extractive Instruction yields effective results, ensuring manageable inference costs. Details are provided in the Appendix \ref{sec:entail}.

(4) \emph{\underline{R}eview Instruction($I_{\underline{R}}$):} 

The responses of the LLM tend to be diverse under high temperature settings \cite{caccia2018language}. During repeated QA testing, the LLM may change its answers to certain `less confident' questions due to the proximity of predicted probabilities. Consequently, after supplemental training, questions initially answered correctly may be answered incorrectly, thus affecting accuracy.

To reinforce the LLM's memory of the knowledge implied by correctly answered questions, we sample the responses provided correctly by the LLM in the previous round. We process some of these responses in the Normative Instruction format and the rest in the Extractive Instruction format to build the Review Instruction.

The composition of the Review Instruction is defined as:
$$
I_{R} = Normative(\alpha r) + Extractive((1-\alpha)r), \nonumber
$$
where $\alpha \in [0,1]$ represents the proportion of responses in the Normative Instruction format, and $r$ denotes the correct responses from the previous round. $Normative(\cdot)$ and $Extractive(\cdot)$ convert responses into the Normative and Extractive Instruction formats, respectively. In subsequent experiments, we examine the effect of varying $\alpha$.

(5) \emph{\underline{G}eneral Instruction($I_{\underline{G}}$):} We replay previous IFT data to mitigate catastrophic forgetting when training LLMs on new data \cite{he-etal-2021-analyzing}. When previous IFT data is unclear, we use the IFT data from the ShareGPT dataset\footnote{https://huggingface.co/datasets/Aeala/ShareGPT\_\\Vicuna\_unfiltered} as the General Instruction to preserve the model's general capabilities on other tasks and instruction-following ability.

\textbf{Supplemental Training Set Construction Strategy. }
We propose a novel CL dataset construction paradigm that parallelizes CIT and CPT data. Specifically, we develop three supplemental training set construction strategies—Plain, Extractive, and Review—by concatenating some or all of five types of data, as shown in Step 5 of Figure \ref{fig:pipeline}. These strategies are named to reflect the specific data components incorporated.




\section{Experimental Setup     }
\subsection{Training Setup}
We train the models on 2 A800 GPUs using ZeRO~\citep{20-zero} stage 1 from DeepSpeed~\citep{20-deepspeed}. 
We adopt AdamW~\citep{14-adamw} as the optimizer and set the batch size to 32 with a maximum sequence length of 2,048 and 1 training epoch. 

We explore the impact of full fine-tuning, LoRA fine-tuning \cite{hu2021lora}, and single-stage versus multi-stage training methods. For details, see the Appendix \ref{app-ft}.

\subsection{Datasets}
We conduct experiments on two question-answering datasets, \textbf{Xiezhi} \cite{gu2023xiezhi} and \textbf{CMMLU} \cite{li2024cmmlu}. To evaluate CEM’s out-of-domain generalization, we apply CEM on Xiezhi as the in-domain dataset and use CMMLU and \textbf{C-Eval} \cite{huang2023ceval} as out-of-domain benchmarks.

To investigate the impact of catastrophic forgetting, we also use the \textbf{HotpotQA} \cite{18-hotpotqa} and \textbf{GSM8K} \cite{21-gsm8k} datasets to test the model's decline in reasoning and mathematical abilities.


\subsection{Base Language Models}


To evaluate the effectiveness of the CEM method in enhancing the performance of LLMs, particularly for smaller parameter-scale models, we select open-source LLMs, including \textbf{Qwen1.5-7B-Chat}\footnote{https://huggingface.co/Qwen/Qwen1.5-7B-Chat} \cite{bai2023qwen}, \textbf{Llama3-8B-Instruct}\footnote{https://huggingface.co/meta-llama/Meta-Llama-3-8B-Instruct} \cite{llama3modelcard}, and \textbf{CuteGPT-13B-ift}\footnote{https://huggingface.co/XuYipei/kw-cutegpt-13b-ift} \cite{CuteGPT}. Each model has been fine-tuned on the IFT data to ensure it effectively follows instructions.

\subsection{Metrics}
The effectiveness of supplemental training will be assessed using the following metrics in the main experiments:

\textbf{ACC}: it represents the accuracy of the answer of LLM.

\textbf{W2R}: it indicates the percentage of questions where the LLM changes incorrect answers to correct ones on the test set.

\textbf{R2W}: it indicates the percentage of questions where the LLM changes correct answers to incorrect ones on the test set.

To analyze the potential performance degradation caused by catastrophic forgetting, we also use the following metrics:

\textbf{ER}: The Enhancement Rate measures the relative improvement brought by CEM on a specific task compared to the initial performance. It is calculated by $ER = \frac{A_k^* - A_k^{0}}{A_k^{0}},$ where $k$ denotes the task using CEM,  $A^0_k$ is the initial accuracy, and $A^*_k$ is the accuracy after CEM.

\textbf{AFR}: The Average Forgetting Rate \cite{wang2024inscl} measures the average relative degradation brought by CEM on tasks, excluding the task using CEM, compared to their initial performance. It is calculated by $AFR = \frac{1}{N-1} \sum_{\substack{i=1 \\ i \neq k}}^{N} \left( \frac{A_i^0 - A_i^*}{A_i^0} \right)$, where $A^0_i$ is the initial accuracy of task $i$ after the initial fine-tuning, and $A_i^*$ is the accuracy of tasks $i$ after applying CEM on task $k$.
\section{Experiments}

We construct supplemental training sets and control groups for experiments as outlined in Section \ref{sec:traningdataconstructionmethods}. To distinguish models trained on different supplemental sets, we use the following naming conventions:

$\textbf{INIT}$ represents the model pre-finetuned solely with the Normative Instruction, equipping it with the ability to generate standardized responses to multiple-choice questions. All subsequent supplemental training is conducted based on this model.

$\textbf{CEM-P}$: \textit{P} for \underline{P}lain, referring to the model trained using the supplemental set constructed with the Plain Strategy.

$\textbf{CEM-E}$ \textbf{and} $\textbf{CEM-E}'$: \textit{E} for \underline{E}xtractive, indicating the model trained using the Extractive Strategy. \textit{E$'$} denotes the ablation group where the General Instruction replaces the Extractive Instruction.

$\textbf{CEM-R}_{\boldsymbol{\alpha}\textbf{=0,0.5,1}}$ \textbf{and} $\textbf{CEM-R}'$: \textit{R} represents the model trained with the \underline{R}eview Strategy supplemental training set. To analyze the impact of the proportion factor $\alpha$ (set to $[0, 0.5, 1]$), we train models $[R_0, R_{0.5}, R_1]$. \textit{$R'$} serves as the ablation group, replacing the Review Instruction with the General Instruction.

We also consider several baselines for comparison with our algorithm as follows:

$\textbf{TempWiki}$: TemporalWiki uses Wikipedia snapshots as CPT data \cite{jang-etal-2022-temporalwiki}. We construct the supplemental training set using corpus from TemporalWiki\footnote{https://huggingface.co/datasets/seonghyeonye/TemporalWiki}, broadly associated with the evaluation set rather than focusing solely on error cases.

$\textbf{AdaptLLM}$: AdaptLLM continues to train the LLMs on the reading comprehension texts constructed based on the raw corpora, mixed with general instructions\cite{cheng2024adapting}.

$\textbf{InstructPT}$: Instruction Pre-Training utilizes an instruction synthesizer\footnote{https://huggingface.co/instruction-pretrain} to convert raw texts from pre-training corpora into instruction-response pairs for continual tuning \cite{cheng2024instructionpretraininglanguagemodels}.

\subsection{Main Results}
We apply the CEM method on the Xiezhi task, collecting approximately 60,000 supplemental corpus samples sourced from Wikipedia. From this corpus, we downsample 25,000 samples as CPT data, while the remaining four types of CIT data are each set to 2,000 samples. To ensure fair comparisons, we add 2,000 Normative Instructions to each of the other baselines, keeping the quantity of raw corpus data consistent. Table \ref{tab:mainres} presents the experimental results, with CMMLU and C-Eval serving as out-of-domain (OOD) benchmarks to evaluate the model’s performance on unseen QA tasks. Our observations are summarized as follows:

\setlength\tabcolsep{3pt} 
\begin{table*}[htp]
\small
  \centering
\begin{tabular}{@{}cccccccc@{}} 
\toprule 
\toprule 
\textbf{Benchmark} & \textbf{INIT}   & \multicolumn{3}{c}{\textbf{Ours}}                & \multicolumn{3}{c}{\textbf{Others}}                                                                             \\ 
\cmidrule(r{2pt}){1-1} \cmidrule(lr{2pt}){2-2} \cmidrule(lr{2pt}){3-5} \cmidrule(l{2pt}){6-8}
\textbf{Acc(\%)}     & \textbf{Prompt} & \textbf{CEM-P} & \textbf{CEM-E} & \textbf{CEM-R\textsubscript{1}} & \textbf{TempWiki} & \textbf{AdaptLLM} & \textbf{InstructPT} \\ \midrule
\rowcolor[rgb]{ .949,  .953,  .961}\multicolumn{8}{c}{\textit{Qwen1.5-7B-Chat}}     \vspace{0.1cm}     \\
Xiezhi             & 42.41           & 51.52          & \textbf{52.12}          & \textcolor{deepred}{\textbf{56.87}} & 30.15                  & 35.14             & 49.05               \\ \midrule
\textcolor{deepgreen}{CMMLU}              & 36.08           & 36.30          & 37.44          & \textcolor{deepred}{\textbf{40.34}} & 24.22                  & 21.12             & \textbf{38.50}               \\ \midrule
\textcolor{deepgreen}{C-Eval}             & 35.00           & 37.30          & \textbf{40.34}          & \textcolor{deepred}{\textbf{40.90}} & 27.10                  & 16.50             & 36.80               \\ \midrule

\rowcolor[rgb]{ .949,  .953,  .961}\multicolumn{8}{c}{\textit{Llama3-8B-Instruct}}   \vspace{0.1cm}   \\                                                                                                                                                                    
Xiezhi             & 25.09           & 35.98          & 40.52          & \textcolor{deepred}{\textbf{54.72}} & 32.24                  & 41.63             & \textbf{42.46}               \\ \midrule
\textcolor{deepgreen}{CMMLU}               & 23.00           & 24.92          & 29.76          & \textcolor{deepred}{\textbf{33.92}} & 23.40                  & 27.62             & \textbf{31.30}               \\ \midrule
\textcolor{deepgreen}{C-Eval}              & 26.00           & 26.20          & 30.20          & \textbf{31.40}          & 22.90                  & 25.30             & \textcolor{deepred}{\textbf{32.50}}      \\ \midrule
\rowcolor[rgb]{ .949,  .953,  .961}\multicolumn{8}{c}{\textit{CuteGPT-13B-ift}}     \vspace{0.1cm}                                                                                                                                                                                                       \\
Xiezhi             & 45.70           & 56.82          & \textbf{57.03}          & \textcolor{deepred}{\textbf{58.67}} & 50.80                  & 45.67             & 54.04               \\ \midrule
\textcolor{deepgreen}{CMMLU}              & 32.56           & 37.96          & \textbf{37.42}          & \textcolor{deepred}{\textbf{38.26}} & 33.44                  & 32.32             & 35.22               \\ \midrule
\textcolor{deepgreen}{C-Eval}              & 31.50           & 35.00          & \textbf{35.20}          & \textcolor{deepred}{\textbf{36.10}} & 31.60                  & 28.50             & 33.00               \\ \bottomrule\bottomrule 
\end{tabular}
 \caption{\label{tab:mainres} The main experimental results of our methods and baseline approaches across various tasks are presented. Experiments are conducted using three different LLMs: Qwen1.5-7B-Chat, Llama3-8B-Instruct and CuteGPT-13B-ift. The top two performances are highlighted in \textcolor{deepred}{\textbf{red bold}} and \textbf{black bold}, respectively.}
  \label{tab:accents}
\end{table*}

\textbf{Firstly, CEM leads to significant accuracy improvements across all models and tasks.} All variations of CEM substantially enhance model performance in both the in-domain (ID) Xiezhi task and OOD tasks. For instance, the Llama3-8B-Instruct model achieves accuracy gains of up to 29.63\% on the Xiezhi task, 10.92\% on the CMMLU task, and 5.40\% on the C-Eval task when using CEM-R. This highlights the effectiveness of the CEM method in addressing the model’s knowledge gaps and its capacity to generalize this knowledge to unseen QA tasks.

\textbf{Secondly, CEM demonstrates superior stability compared to other baselines.} Both TempWiki and AdaptLLM occasionally result in performance degradation for LLMs. For instance, AdaptLLM reduces the accuracy of the Qwen1.5-7B-Chat model by 7.27\% on the Xiezhi task, 14.96\% on the CMMLU task, and 18.50\% on the C-Eval task. In contrast, all variations of CEM consistently deliver positive performance improvements.

\textbf{Thirdly, Extractive Instruction and Review Instruction further enhance performance.} A clear trend of incremental performance improvement is observed from CEM-P to CEM-E and finally to CEM-R. The ablative properties introduced during dataset construction suggest that these two types of CIT data play a crucial role in driving the observed improvements.

\subsection{Data Source Analysis}
We analyze the impact of the Supplementary Corpus from different sources—Wikipedia, Bing, and their concatenation—on CEM’s performance. To further validate CEM’s applicability, CMMLU is included as an ID task scenario with data proportions consistent with the main experiment.

\begin{figure}[htb]
  \centering
  \includegraphics[width=0.5\textwidth]{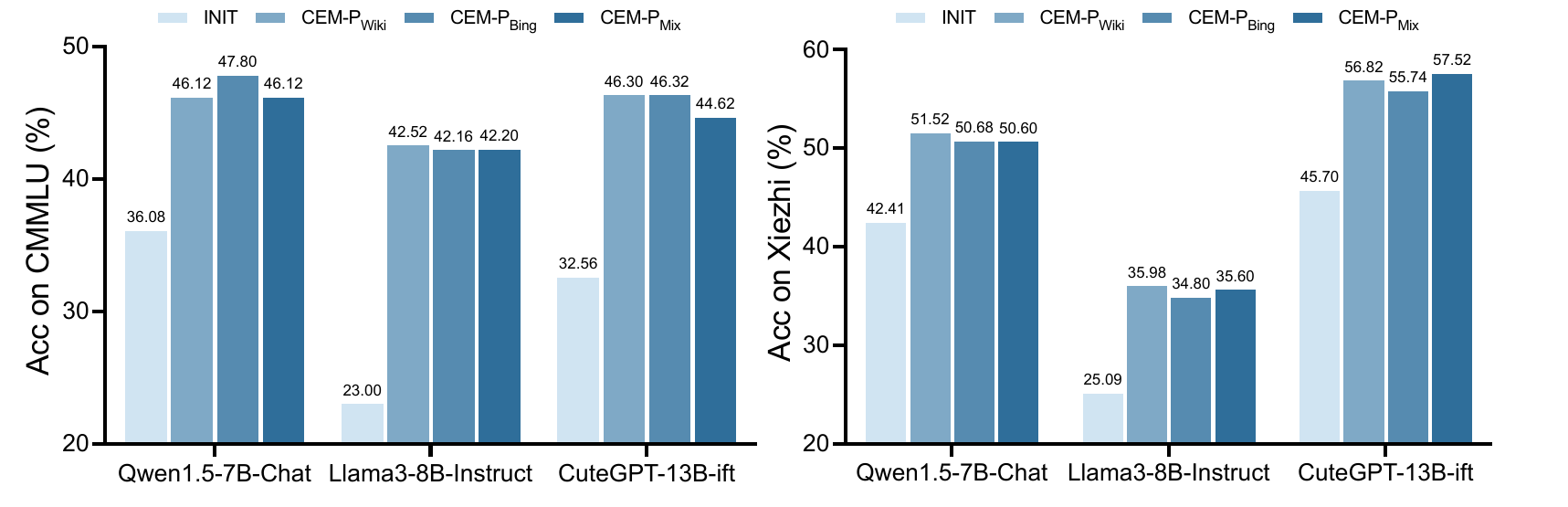}
  \caption{This figure presents the accuracy of models using the CEM-P method with different data sources on Xiezhi and CMMLU tasks. The suffixes \textit{Wiki}, \textit{Bing}, and \textit{Mix} indicate the sources of the Supplementary Corpus, with \textit{Mix} representing using double the samples sourced from the combined dataset of Wikipedia and Bing.}
  \label{fig:chat1}
\end{figure}

We find that \textbf{the CEM method shows robust compatibility across different data sources, consistently improving model performance.} However, integrating multiple sources does not necessarily outperform using a single source. As shown in Figure \ref{fig:chat1}, no significant differences in CEM effectiveness emerge when employing Wikipedia, Bing, or their combination. Although the \textit{Mix} dataset is larger, it often yields lower accuracy than \textit{Wiki} or \textit{Bing}, potentially due to data redundancies. Moreover, an excessively large supplemental corpus can dilute the influence of CIT data, reducing its capacity to mitigate catastrophic forgetting, enhance knowledge extraction, and prevent overfitting.

\subsection{Mechanistic Analysis of Extractive and Review Instruction Components} 
\begin{figure}[htp]
  \centering
  \includegraphics[width=0.5\textwidth]{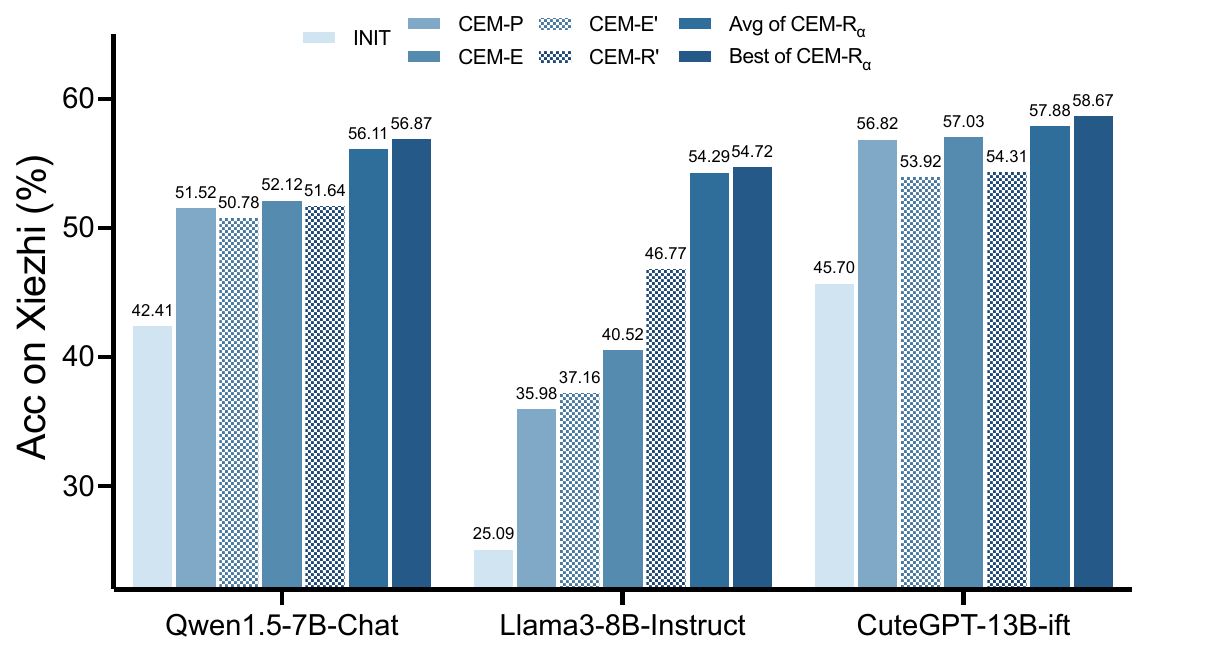}
  \caption{ The figure shows the ablation experimental results of three different models on the Xiezhi task after training with CEM. `Avg of CEM-R$\alpha$' and `Best of CEM-R$\alpha$' represent the average and best results of the CEM-R with $\alpha$ of [0, 0.5, 1].}
  \label{fig:chat2}
\end{figure}

\begin{figure}[htp]
  \centering
  \includegraphics[width=0.5\textwidth]{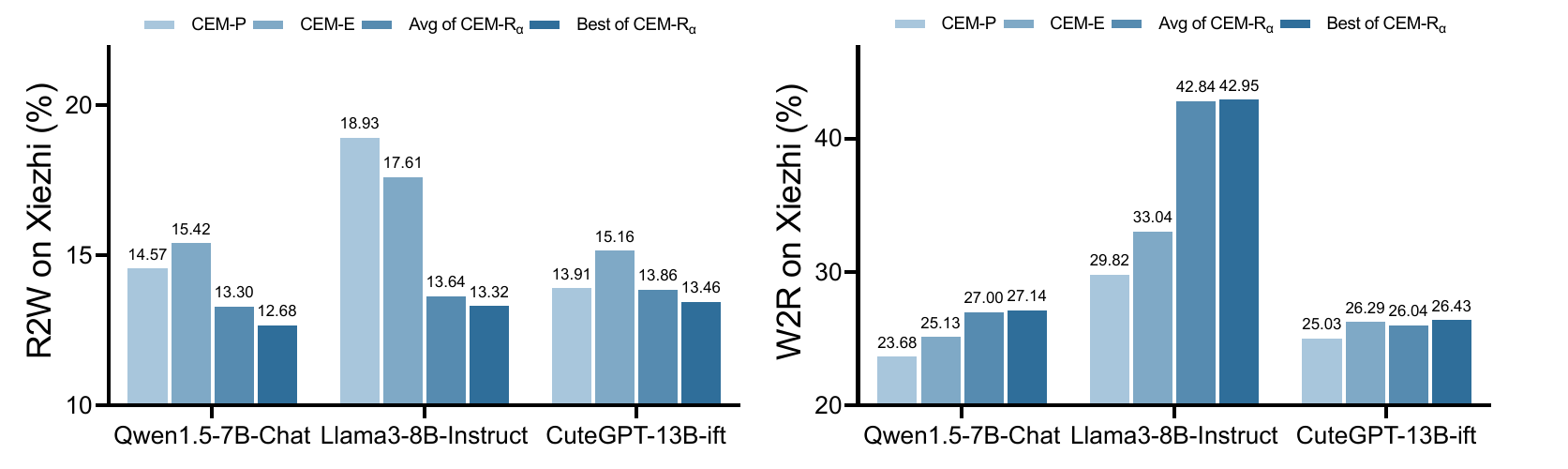}
  \caption{The figure presents the R2W and W2R metrics of the models after CEM supplemental training. `Avg of CEM-R$\alpha$' and `Best of CEM-R$\alpha$' represent the average and best results of the CEM-R with $\alpha$ of [0, 0.5, 1].}
  \label{fig:rw}
\end{figure}

To clarify the underlying mechanisms by which Extractive Instruction and Review Instruction affect model performance, we conduct experiments on CuteGPT-13B-ift and examine the R2W and W2R metrics. Additionally, we explore the effects of varying $\alpha$ (with values $[0, 0.5, 1]$) on CEM-R to determine whether more diverse and complex Review Instructions yield improved results. The corresponding results are illustrated in Figure \ref{fig:chat2} and Figure \ref{fig:rw}.

We summarize the key observations and insights derived from these experiments:

(1) CEM-E and CEM-R outperform their ablation groups with General Instruction, demonstrating the effectiveness and positive impact of Extractive and Review Instruction on model performance. As shown in Figure \ref{fig:rw}, the Extractive Instruction enhances W2R performance: compared to CEM-P, CEM-E achieves improvements of 1.45\%, 3.22\%, and 1.26\% across the three models, indicating more effective error correction. In addition, the Review Instruction leads to a significant reduction in the R2W metric, indicating a reduced tendency for the model to forget previously acquired correct knowledge. For example, using CEM-R\textsubscript{1}, Llama3-8B-Instruct’s R2W performance decreases by 5.61\% and 4.29\%, respectively compared to CEM-P and CEM-E.

While the performance benefits of the Review Instruction are readily understandable, the exact mechanism by which the Extractive Instruction improves the model’s ability to integrate crucial knowledge from lengthy passages warrants further discussion.

We propose the following explanation: The Extractive Instruction provides the model with both the questions and supplementary materials containing relevant information. Questions and answer options typically present key points in a concise manner, whereas the supplementary materials often contain a dense network of factual statements \cite{jiang2024instructiontuned}. Thus, the questions act as a structured induction of the supplementary materials, explicitly linking complex information to the targeted induction (\textit{i.e.,} the questions and options). Serving as cues and guides, the questions and options help the model retrieve pertinent information from the corpus while filtering out redundant or irrelevant details. This process enhances the model’s ability to efficiently extract and utilize essential information.

(2) The proportion factor $\boldsymbol{\alpha}$ has minimal impact on the effectiveness of Review Instruction, as R\textsubscript{0}, R\textsubscript{0.5}, and R\textsubscript{1} exhibit similar effects on LLM performance. This indicates that reviewing knowledge, regardless of format, effectively consolidates it.


\subsection{Multiple Iterations Analysis}

To evaluate the iterability of the CEM method and its potential for catastrophic forgetting during multiple training rounds, we conduct three rounds of CEM training on CuteGPT-13B-ift for the Xiezhi task, using consistent training strategies and non-repetitive CIT data of equivalent size in each round. Specifically, the CEM method is applied to the Xiezhi task, while HotpotQA and GSM8K are used to assess reasoning and mathematical skills. The ER metric evaluates performance improvements on Xiezhi, and the AFR metric monitors performance degradation in other tasks. Initially, 2,000 instances from each dataset are sampled for pre-finetuning to establish a baseline. To address forgetting, we implement a random replay strategy, replacing General Instructions with 400 randomly replayed instances from HotpotQA and GSM8K in each training round.

\begin{figure}[htb]
  \centering
  \includegraphics[width=0.5\textwidth]{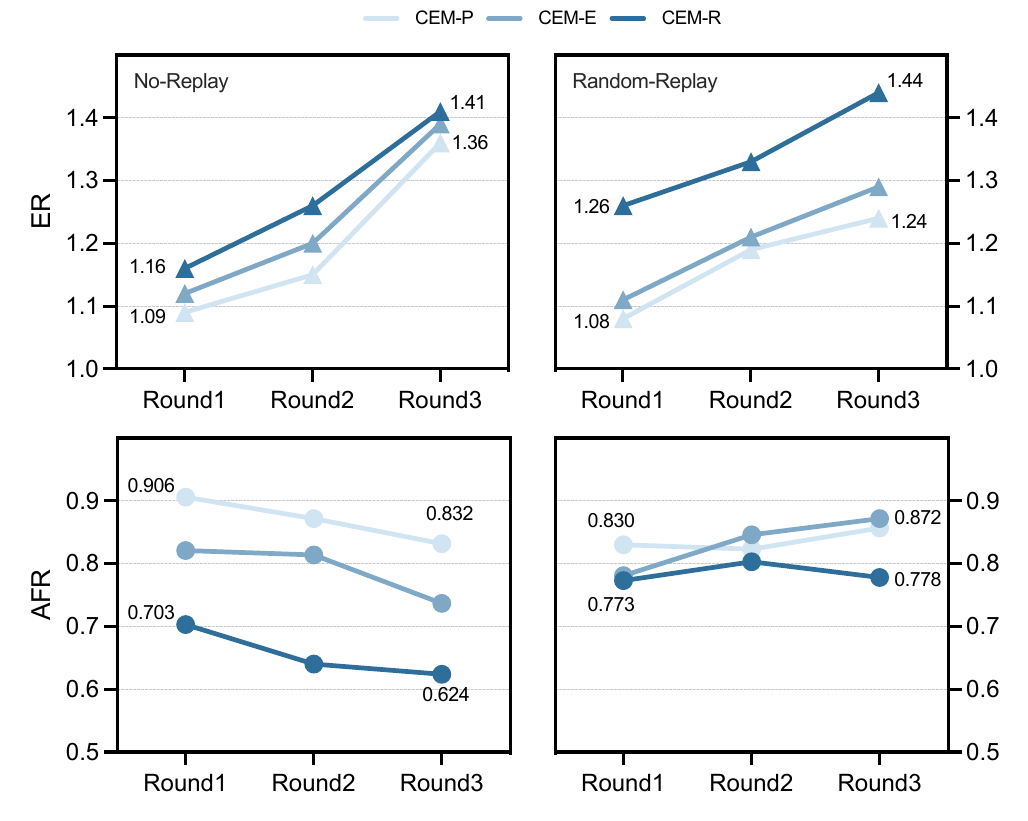}
  \caption{ The figure shows the experimental results of CuteGPT-13B-ift on the Xiezhi task after multiple rounds of CEM. The left part represents results without replay, while the right part shows results with a random replay strategy. ER indicates the enhancement rate on the Xiezhi task, and AFR represents the average forgetting rate on the HotpotQA and GSM8K tasks.  }
  \label{fig:forgetting}
\end{figure}

Figure \ref{fig:forgetting} shows that even after three rounds of CEM, performance on Xiezhi improves significantly and maintains an upward trend. The combination of CEM and random replay effectively mitigates performance decline, with AFR exceeding 75.0\% in most cases, highlighting its iterative potential. However, without a replay strategy, initial rounds show better improvement (possibly because the model does not need to adapt to the more complex multitasking instructions introduced by random replay), but more rounds lead to greater forgetting, limiting further iteration, as seen when CEM-R reaches an ER of 62.4\% in round three without replay. CEM-R, with its larger data capacity, achieves significant performance improvements but shows the greatest decline in other tasks, likely due to the increased volume of CPT data diluting the replay effect. Combining CEM with other advanced replay-based methods and data balancing warrants further investigation.

\section{Conclusion}


In this paper, we propose the Continue Evolving from Mistakes (CEM) method, a data-efficient and cost-effective method for collecting CPT data and continually evolving LLMs on specific tasks by identifying and addressing internal knowledge deficiencies revealed by models' mistakes. CEM employs a pipeline that leverages model mistakes to gather extensive and targeted CPT data, combined with a novel parallel training set construction paradigm integrating both CIT and CPT data for continuous training. Experiments on various open-source LLMs show that CEM significantly enhances models' answer accuracy and supplements their internal knowledge deficiencies. These improvements effectively generalize to out-of-domain tasks. Furthermore, when integrated with replay-based methods, CEM demonstrates substantial potential for sustained evolution across multiple iterations while mitigating performance degradation.

\section{Limitations} 
This paper focuses on the model's performance in question-answering tasks. It remains uncertain whether CEM can consistently enhance the model's capabilities and performance in non-QA tasks.

Future research will focus on investigating additional rounds of supplemental training to further enhance the LLM and explore the integration of other strategies to mitigate catastrophic forgetting. Additionally, it will examine whether repeated supplemental learning on already mastered corpora could have adverse effects.

Furthermore, we will experiment with diverse data sources, including synthetic corpora, and optimize strategies for collecting supplementary corpora to endow the LLM with advanced continual iterative evolutionary capabilities.

\bibliography{custom}
\appendix
\section{Knowledge Point Extraction}
\label{appendix:kp}
Table \ref{tab:labeling} shows the prompt used for extracting knowledge points from questions using Qwen2-7B-Instruct. In the experiment, the variable  $x$  in the prompt is set to $5$.

\begin{table*}[htp]
\centering
\begin{tabularx}{1.0\textwidth} { 
   >{\raggedright\arraybackslash}X 
   }
\hline
\textbf{An Example of Knowledge Point Extraction}\\ 
\hline
\textbf{Instruction:}\\
\textcolor{red}{\emph{(task)}}Please analyze the core knowledge points examined by the following question.\\
\textcolor{red}{\emph{(question)}}\#\#\# Question: \\
What is the main component of the cell membrane? \\
A. Phospholipids  B. Carbohydrates  C. Proteins  D. Nucleus \\
\textcolor{red}{\emph{(answer)}}\#\#\# Answer \\
The answer is A. Phospholipids. Phospholipids are the main component of the cell membrane. \\
\textcolor{red}{\emph{(requirement)}}\#\#\# Requirements: \\
- Prioritize identifying directly relevant named entities from the question and options. \\
- Knowledge points should be closely related to the question and options, aiding in eliminating incorrect options and selecting the correct one. \\
- Knowledge points should be specific, avoiding overly broad, common, or indistinct concepts, e.g., "Prime Minister", "Archaeology".  \\
- Return the knowledge points in a list format, e.g., ["English Civil War", "Glorious Revolution"]. \\
- The number of returned knowledge points should not exceed \{x\}. \\

\hline
\textbf{Output:}\\
\textcolor{red}{\emph{(answer)}}["Cell Membrane Structure", "Phospholipid Bilayer"] \\ 
\hline
\end{tabularx}
\caption{\label{tab:labeling}An example illustrating the extraction of knowledge points from questions.}
\end{table*}

\section{Effectiveness Check of Supplemental Corpus}
\label{sec:relevant}

Table \ref{tab:entail} shows the prompt used for entailment checking of the Supplemental Corpus with Qwen2-72B-Instruct. The results are presented in Table \ref{tab:useful}, where the proportion of useful information from both data sources consistently exceeds 65\%.

\begin{table}[H]
  \centering
  \begin{tabular}{ccc}
    \hline
    \textbf{Data Source} & \textbf{Data Volume} & \textbf{Effectiveness} \\
    \hline
    Bing & 8161 &  0.6718\\
    Wiki & 7694 &  0.6508 \\
    \hline
  \end{tabular}
  \caption{\label{tab:useful} The results of verifying whether the Supplemental Corpus contains information that helps the LLM in correcting errors and answering questions accurately. `Effectiveness' denotes the proportion of useful information.}
\end{table}

\section{Process of Collecting Corpora from Multiple Sources}
\label{appendix:wikibing}
The parameter configuration for the experiment is as follows: the value of $k$ is set to $4$ for top-k retrieval. Indexing is implemented using FAISS’s IndexFlatIP. The threshold for inner product similarity is set at $0.80$. If no entries meet this threshold, the entry with the highest similarity exceeding $0.70$ is selected; otherwise, no result is returned.

\begin{figure*}[htbp]
  \centering
  \includegraphics[width=0.9\textwidth]{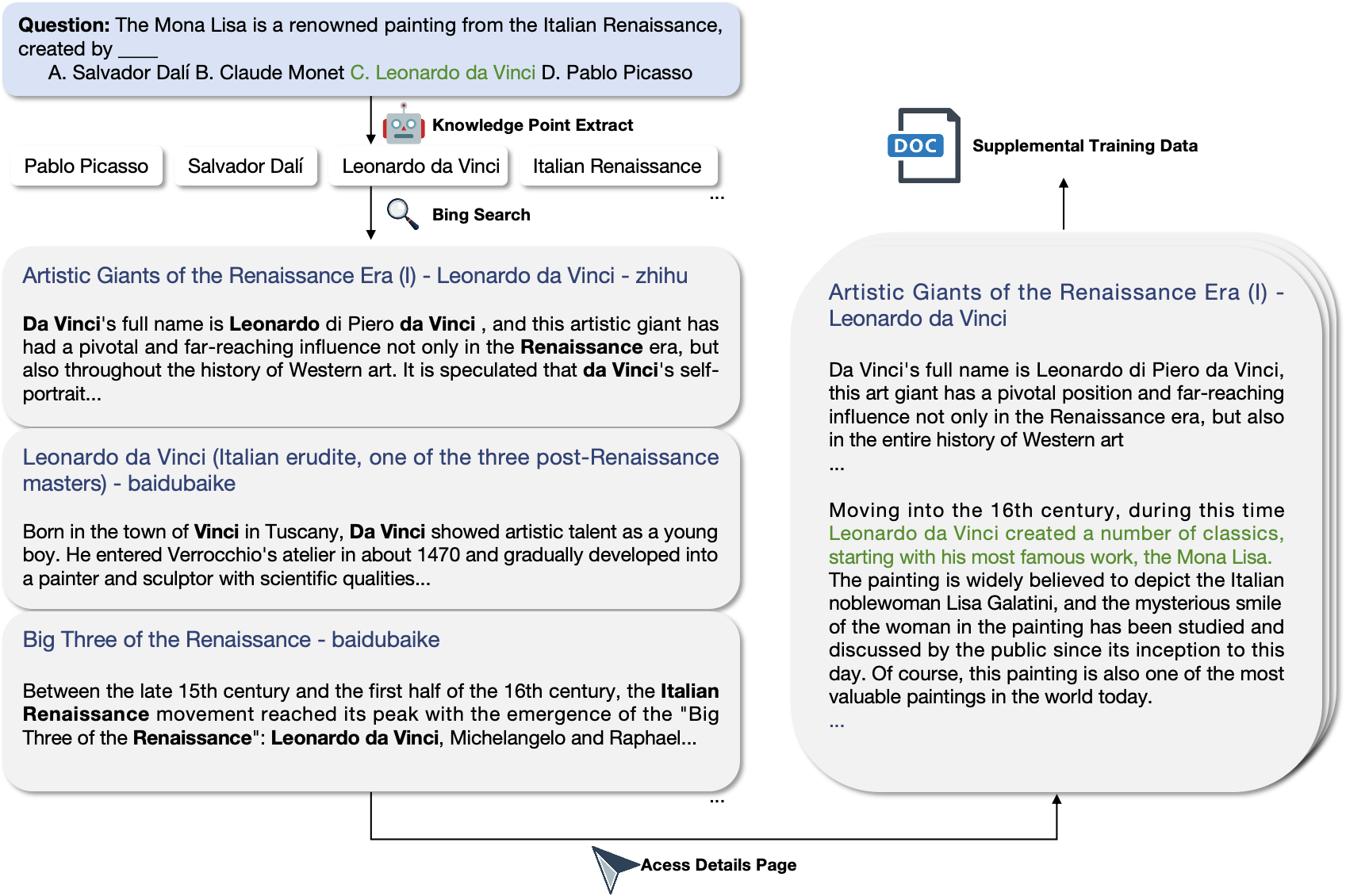}
  \caption{The process of collecting supplemental corpus from Bing.}
  \label{fig:bing}
\end{figure*}

\begin{figure*}[htbp]
  \centering
  \includegraphics[width=0.9\textwidth]{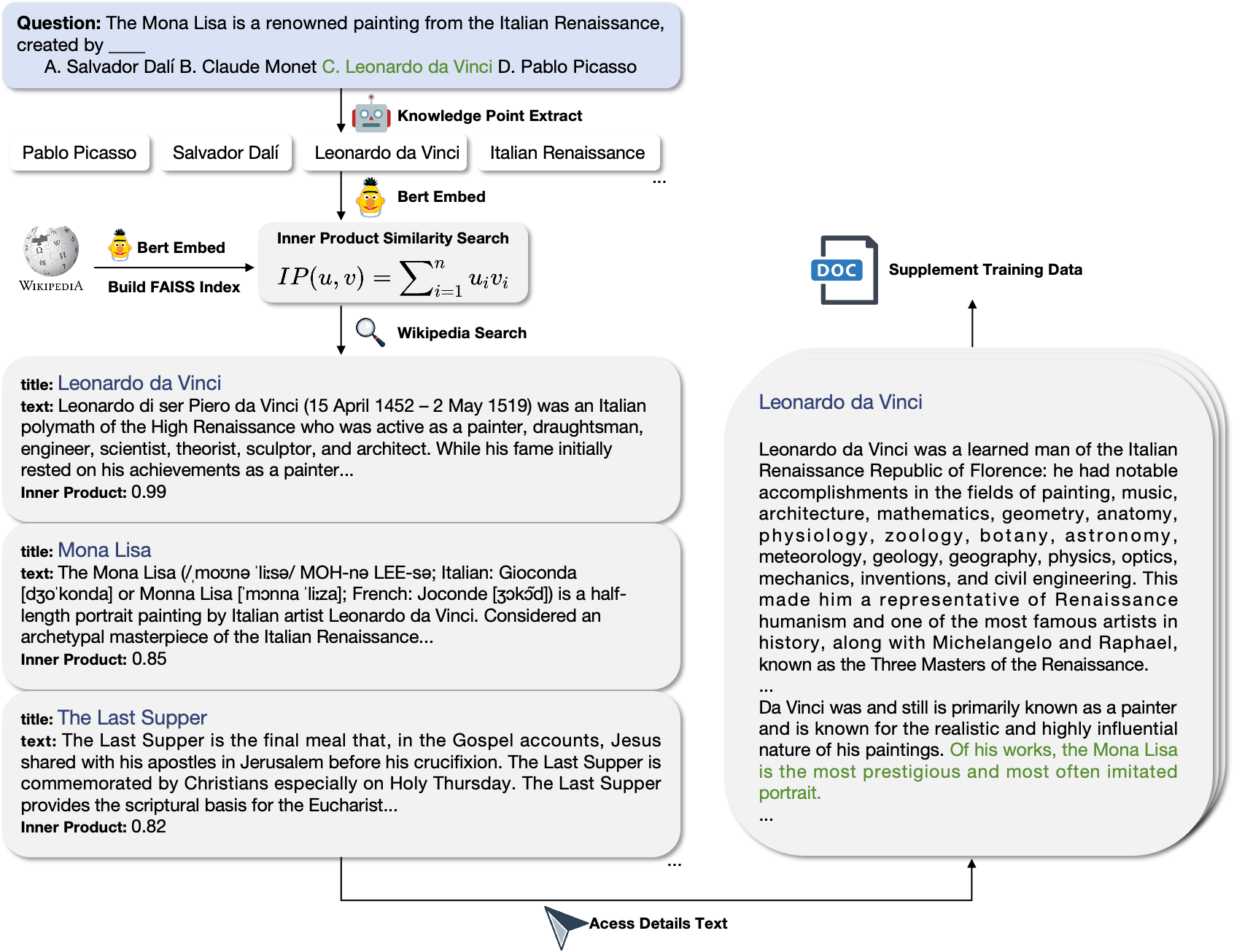}
  \caption{The process of collecting supplemental corpus from Wikipedia.}
  \label{fig:wiki}
\end{figure*}

Fig.\ref{fig:bing} and Fig.\ref{fig:wiki} show the process of obtaining supplemental data from Bing and Wikipedia respectively.

\section{Format of CIT Data}
\label{sec:cit}
Table \ref{tab:normative} and Table \ref{tab:extractive} show the IFT data formats of the Normative Instruction and the Extractive Instruction.

\begin{table*}[htp]
\centering
\begin{tabularx}{1.0\textwidth} { 
   >{\raggedright\arraybackslash}X 
   }
\hline
\textbf{An Example of Normative Instruction}\\ 
\hline
\textbf{Instruction:}\\
\textcolor{red}{\emph{(task)}}Please select the correct answer for the following single choice questions:\\
\begin{tabular}[c]{@{}l@{}}\textcolor{red}{\emph{(question)}}\#\#\# Question: \\
In the 2nd century, what became the Roman Empire's internal sea was \_\_\_\_\_\\ A. Red Sea  B. Caspian Sea  C. Mediterranean Sea  D. Black Sea\end{tabular} \\ \hline
\textbf{Output:}\\
\textcolor{red}{\emph{(answer)}}The answer is C. Mediterranean Sea. In the 2nd century, what became the Roman Empire's internal sea was Mediterranean Sea.\\ 
\hline
\end{tabularx}
\caption{\label{tab:normative}An example illustrating the format of the Normative Instruction.}
\end{table*}

\begin{table*}[htp]
\centering
\begin{tabularx}{1.0\textwidth} { 
   >{\raggedright\arraybackslash}X 
   }
\hline
\textbf{An Example of Extractive Instruction}\\ 
\hline
\textbf{Instruction:}\\
\textcolor{red}{\emph{(task)}}Please select the correct answer for the following single choice questions based on the supplementary materials provided:\\
\textcolor{red}{\emph{(question)}}\#\#\# Question: \\
In the 2nd century, what became the Roman Empire's internal sea was \_\_\_\_\_\\ A. Red Sea  B. Caspian Sea  C. Mediterranean Sea  D. Black Sea \\
\textcolor{red}{\emph{(supplement)}}\#\#\# Supplementary Materials: \\
The Roman Empire ... During the reign of Tulajin (98-117), the Roman Empire reached its greatest extent with the Mediterranean Sea becoming the empire's inland sea. In its heyday, it controlled about 5 million square kilometers of land, making it one of the largest monarchies in the ancient history. \\ 
\hline
\textbf{Output:}\\
\textcolor{red}{\emph{(answer)}}The answer is C. Mediterranean Sea. In the 2nd century, what became the Roman Empire's internal sea was Mediterranean Sea. \\ 
\hline
\end{tabularx}
\caption{\label{tab:extractive}An example illustrating the format of the Extractive Instruction. The ellipsis indicates that some supplemental text has been omitted for clarity.}
\end{table*}

\section{Entailment Checking of Extractive Instruction}
\label{sec:entail}

Table \ref{tab:entail} shows the prompt used for entailment checking of the Extractive Instruction with Qwen2-72B-Instruct.

\begin{table*}[htp]
\centering
\begin{tabularx}{1.0\textwidth} { 
   >{\raggedright\arraybackslash}X 
   }
\hline
\textbf{An Example of Entailment Checking}\\ 
\hline
\textbf{Instruction:}\\
\textcolor{red}{\emph{(task)}}Can the correct answer to the given multiple-choice question be derived solely from the knowledge in the following material? Please respond with 'Yes' or 'No'. \\
\textcolor{red}{\emph{(question)}}\#\#\# Question: \\
What is the main component of the cell membrane? \\
A. Phospholipids  B. Carbohydrates  C. Proteins  D. Nucleus \\
\textcolor{red}{\emph{(answer)}}\#\#\# Answer \\
The answer is A. Phospholipids. Phospholipids are the main component of the cell membrane. \\
\textcolor{red}{\emph{(supplement)}}\#\#\# Materials: \\
The phospholipid bilayer is a thin polar membrane made of two layers of lipid molecules. These membranes are flat sheets that form a continuous barrier around all cells. The cell membranes of almost all organisms and many viruses are made of a lipid bilayer, as are the nuclear membrane surrounding the cell nucleus, and membranes of the membrane-bound organelles in the cell... \\

\hline
\textbf{Output:}\\
\textcolor{red}{\emph{(answer)}} Yes. \\ 
\hline
\end{tabularx}
\caption{\label{tab:entail}An example illustrating the entailment checking of the Extractive Instruction. The ellipsis indicates that some supplemental text has been omitted for clarity.}
\end{table*}

\section{Exploring Fine-Tuning Strategies}
\label{app-ft}
We experiment with fine-tuning strategies on Qwen-7B-Chat, employing full fine-tuning and LoRA fine-tuning for parameter updates. For dataset usage, we apply two approaches: single-stage parallel training on combined CPT and CIT data (3000 and 2000 samples, respectively), and multi-stage sequential training, where the model is first fine-tuned on CPT data, followed by further fine-tuning on CIT data. 

Table \ref{tab:ft} shows that parallel full fine-tuning yields the best results. Therefore, in the experiments, except for ChatGLM-6B, which employs Prompt-tuning \cite{liu2022ptuning}, all other models use parallel full fine-tuning.

\begin{table}[H]
  \centering
  \begin{tabular}{cccc}
    \hline
      & \textbf{Pre-Finetuning} & \textbf{Parallel} & \textbf{Sequential} \\
    \hline
    \textbf{Full} &0.4162 & 0.4854 & 0.3154 \\
    \textbf{LoRA} &- & 0.4573 & 0.4271 \\
    \hline
  \end{tabular}
  \caption{\label{tab:ft} Results of verifying the impact of CPT and CIT datasets on model performance using different parameter update strategies and training stages.}
\end{table}

\end{document}